%% file: templateArxiv.tex
\documentclass{article}

\usepackage{PRIMEarxiv}
\usepackage{amsmath,amsfonts}
\usepackage[utf8]{inputenc} 
\usepackage[T1]{fontenc}    
\usepackage{hyperref}       
\usepackage{url}            
\usepackage{booktabs}       
\usepackage{amsfonts}       
\usepackage{nicefrac}       
\usepackage{microtype}      
\usepackage{lipsum}
\usepackage{fancyhdr}       
\usepackage{graphicx}       
\graphicspath{{media/}}     

\title{Adaptive Robot Localization with Ultra-wideband Novelty Detection}

\author{
  Umberto Albertin\\
  Department of Electronics \\
  and Telecommunications (DET) \\
  Politecnico di Torino \\
  Torino, TO 10129\\
  \texttt{umberto.albertin@polito.it} \\
    \And
  Mauro Martini\\
  Department of Electronics and \\
  Telecommunications (DET) \\
  Politecnico di Torino \\
  Torino, TO 10129\\
  \texttt{mauro.martini@polito.it} \\
  \And
  Alessandro Navone\\
  Department of Electronics and \\
  Telecommunications (DET) \\
  Politecnico di Torino \\
  Torino, TO 10129\\
  \texttt{alessandro.navone@polito.it} \\
  \And
  Marcello Chiaberge\\
  Department of Electronics and \\
  Telecommunications (DET) \\
  Politecnico di Torino \\
  Torino, TO 10129\\
  \texttt{marcello.chiaberge@polito.it}
}

\begin{document}
\maketitle

\begin{abstract}
Ultra-wideband (UWB) technology has shown remarkable potential as a low-cost general solution for robot localization. However, limitations of the UWB signal for precise positioning arise from the disturbances caused by the environment itself, due to reflectance, multi-path effect, and Non-Line-of-Sight (NLOS) conditions. This problem is emphasized in cluttered indoor spaces where service robotic platforms usually operate.
Both model-based and learning-based methods are currently under investigation to precisely predict the UWB error patterns. Despite the great capability in approximating strong non-linearity, learning-based methods often do not consider environmental factors and require data collection and re-training for unseen data distributions, making them not practically feasible on a large scale.
The goal of this research is to develop a robust and adaptive UWB localization method for indoor confined spaces. A novelty detection technique is used to recognize outlier conditions from nominal UWB range data with a semi-supervised autoencoder. Then, the obtained novelty scores are combined with an Extended Kalman filter, leveraging a dynamic estimation of covariance and bias error for each range measurement received from the UWB anchors. The resulting solution is a compact, flexible, and robust system which enables the localization system to adapt the trustworthiness of UWB data spatially and temporally in the environment. The extensive experimentation conducted with a real robot in a wide range of testing scenarios demonstrates the advantages and benefits of the proposed solution in indoor cluttered spaces presenting NLoS conditions, reaching an average improvement of almost $60 \%$ and greater than $25~cm$ of absolute positioning error.

\textit{Note to Practitioners} 
--- UWB technology is widely used for robot localization due to its capability to provide quite precise distance measurements between a tag and an anchor. These distance measurements are used in localization algorithms such as the Kalman filter, which are essential for accurately localizing the robot during the motion.
The motivation for this research stems from robot localization challenges within indoor spaces, such as private environments or industrial facilities. Such environments introduce signal interference from walls or metal furniture, leading to distortions in UWB distance measurements and generating localization errors. This paper introduces a novel framework aimed at mitigating the impact of environmental noise on localization. The proposed framework consists of two main components: firstly, a novelty detection neural network is employed to assess deviations from nominal training conditions due to environmental factors, such as new obstacles or unseen external features. Secondly, an Extended Kalman Filter uses the output of the neural network as a scoring mechanism to adjust the related UWB measurement uncertainty depending on environmental interference. Additionally, this work includes an ablation study to further demonstrate the efficacy of the proposed method compared to existing state-of-the-art solutions. Future works plan to improve the actual framework by studying networks with online training techniques and time-based inputs. 
\end{abstract}


\section{Introduction}
\label{sec:intro}
The rapid emergence of service mobile robots marks a significant advancement in automation, impacting various aspects of daily life, including domestic and health assistance \cite{ eirale2022human, tamantini2021robotic}, precision agriculture \cite{navone2023rows, martini2022}, and facilities inspection \cite{gehring2021anymal}.  Central to the future development and reliability of autonomous mobile robots are localization technologies \cite{alatise2020review}. While wheel odometry and visual odometry have been extensively studied over the past decade, they face challenges with long paths and poor repetitive features \cite{navone2023online, wang2020approaches, gupta2020corridor}.

Among the various localization technologies developed in the last decades, Ultra-Wideband (UWB) has recently gained attention as a promising, cost-effective solution for localizing mobile robots and devices in environments where GPS is unavailable \cite{elsanhoury2022precision}.
UWB technology operates by transmitting extremely short-duration pulses, enabling precise time-of-arrival estimations and highly accurate distance measurements. However, despite its potential, UWB is susceptible to significant sources of uncertainty, such as Non-Line of Sight (NLoS) propagation conditions and multi-path reflections, which can substantially degrade localization accuracy \cite{gezici2005localization}. These issues arise when the signal cannot travel directly from the transmitter to the receiver or when it reflects off surfaces, leading to biased distance estimations between devices and hindering autonomous navigation tasks. 

Disparate methods have been proposed in the last years to mitigate such limitations, as deeply discussed in Section \ref{sec:related_works}. Among them, the majority of localization approaches are based on Kalman filtering \cite{feng2020kalman, karfakis2022uwb}. On the other hand, Machine Learning is spreading in data sciences and robotics realms, substituting model-based systems with data-driven approaches. Many trials can be found in the literature with the aim of learning a characterization of UWB reflectance and NLoS disturbance errors \cite{bregar2018improving,che2022feature, angarano2021robust}. Despite the great capability of neural networks to approximate strong non-linearity, the resulting solutions require costly dataset collection and re-training for any change in environmental conditions.
Nonetheless, the dynamic nature of the environments in which service robots operate further complicates the identification of areas where the reliability of the UWB localization system may decline. 

\begin{figure*}[ht]
\centering
\includegraphics[width=\textwidth]{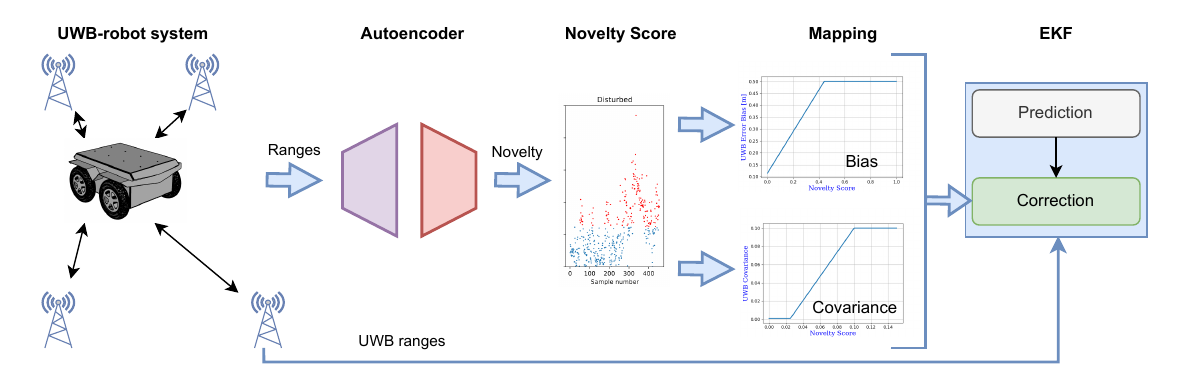}
\caption{The complete pipeline of the proposed adaptive UWB positioning solution. UWB ranges are used to feed a novelty autoencoder, the novelty score is then dynamically mapped into a correction bias and a covariance value for the correction step of the EKF.}
\label{fig:pipeline_schematic}
\end{figure*}

Hence, the core focus of this work is to develop a robust solution to learn a precise characterization of the environment in terms of range signal reliability and exploit it to adapt the UWB localization system over regions and time. This can be achieved by mixing an Extended Kalman Filter (EKF) with the advantages of unsupervised neural network training. In particular, we adopted a novelty detection technique to train an autoencoder network and to identify subtle changes in the UWB ranges. The overall system, schematically described in Figure \ref{fig:pipeline_schematic}, allows the linking of the novelty score representing the uncertainty in the UWB ranges to bias and covariance parameters of the EKF with a simple mapping function while the robot is moving around the environment. We show that to train the novelty autoencoder with nominal conditions, we only need to collect UWB data with a reduced number of trajectories, avoiding a fingerprinting approach previously presented \cite{albertin2024case}.
Moreover, extensive experimentation is conducted to test the positioning performance of the system compared to the static EKF baseline, collecting ground truth trajectories with a real robot in various environmental settings. An ablation study is finally presented, highlighting the impact of key parameters on the results, above all, the number of UWB anchors and the estimated bias correction term.

The main contributions of this work can be, therefore, summarized as:
\begin{itemize}
    \item a novelty detection autoencoder to learn the characterization of the UWB range signals in the environment. The training of the autoencoder only requires a few robot trajectories to collect UWB range data without costly labeling or fingerprint approaches;
    \item a localization system that leverages the range-specific novelty score obtained from the autoencoder to dynamically estimate a covariance and a bias term for EKF correction step with simple and general mapping functions;
    \item an extensive experimentation and ablation of the localization system on nine scenarios with increasing levels of NLoS condition and dynamic evolution of the environment that, to the best of our knowledge, has never presented before.
\end{itemize}

\section{Related works}
\label{sec:related_works}
UWB technology has emerged as a leading solution for precise localization, particularly in complex indoor environments where traditional systems are ineffective. UWB's broad frequency spectrum enables high temporal resolution, facilitating accurate distance measurements and positioning \cite{mazhar2017precise}. This positioning system has become popular due to its accuracy, which can reach the centimeter level, thanks to its elevated resolution \cite{gezici2005localization}.  Moreover, it is immune to inferences from other signals due to its different signal types and radio spectrum and can penetrate different materials, even though it suffers from heavy signal modifications. Commonly, UWB localization systems are composed of a two-step approach. At the beginning, some parameters related to the target node are estimated through some fixed nodes placed in a known position. Among the most common estimated parameters, Time of Arrival (ToA), Angle of Arrival (AoA) and Received Signal Strenght (RSS), cover a major role \cite{farid2013recent}. As a second phase, different algorithms are employed to estimate directly the localization of the target node.

The integration of UWB technology with advanced filtering techniques, such as Kalman filters, has become a focal point in indoor localization research, aiming to improve accuracy and reliability. Studies have shown that combining these approaches with other methodologies significantly enhances location estimation precision in complex indoor settings. In fact, Kalman filter techniques allow to fuse the data of UWB with other sensors, obtaining an improved position estimation. In particular, many examples of sensor fusion can be found in literature, where UWB is integrated with other localization sources, such as inertial sensors \cite{feng2020kalman, corrales2008hybrid}, visual-inertial sensors \cite{benini2013imu}, and GPS-inertial sensors \cite{krishnaveni2022indoor}.

In recent years, there has been a significant rise in the use of machine learning techniques to enhance UWB localization, as extensively discussed in the literature. These techniques are utilized either to correct individual anchor-to-tag range measurements or to refine the final estimated position of the target. For example, various neural network architectures have been explored to mitigate localization errors effectively. Deep autoencoders have been proposed for the direct estimation of the 2D position, exploiting the time difference of arrival \cite{ye2020research}. More complex networks, such as recurrent neural networks, in particular Long-Short Term Memory (LSTM), have been employed for estimating the direct position of the tag leveraging the encoding of temporal dependencies and learning high-level representation from the extracted features \cite{poulose2021feature}. Convolutional neural networks have also been employed for estimating the position of a UWB device, such as in \cite{nguyen2020convolutional}, where an RGB image is generated from the received signal and then given as input to the network.
Instead, many studies focus on algorithms aimed at mitigating errors on single ranges.
For example, deep learning and machine learning methodologies have been proposed to address NLoS conditions \cite{bregar2018improving}, framing the issue as both a classification and regression problem \cite{che2022feature, angarano2021robust}. Other approaches have incorporated additional data sources, such as inertial measurements \cite{goudar2021online} or map-based information \cite{wang2021high}, to enhance the accuracy of UWB localization. In certain instances, receiver positions have been estimated directly from raw data, circumventing the necessity to pre-identify NLoS conditions. These techniques have demonstrated remarkable accuracy even in complex environments where maintaining a line of sight (LoS) condition is not always feasible \cite{nitsoo2019deep}. Nevertheless, a comprehensive solution capable of effectively handling the diverse and dynamic nature of real-world environments remains elusive.

In this study, we highlight how the environment has a central role on the overall performance of the UWB, considering the dynamic nature of UWB error strictly coupled with the introduction of anomalies in the map. To this aim, novelty detection is a set of machine learning techniques capable of identifying new or unknown patterns in data distribution starting from nominal learned data \cite{gruhl2021novelty}. In the context of UWB signals, novelty detection can identify NLoS and multipath conditions as novel situations, as well as significative changes in the environment, providing the necessary position-specific uncertainty information for a reliable localization. 

As we highlight in this work, adapting the values of the measurement covariance of the Extended Kalman Filter can provide a crucial improvement for the localization problem, enabling the integration of information about the environment inside the localization algorithm.
For instance, \cite{karfakis2022uwb} utilizes a neural network to model the relationship between positional errors in UWB data and signal quality metrics, such as precision estimates and received signal strength. Results show that neural networks is able to capture sensor covariance and refine the reliability of EKF estimates. This adaptability allows the networks to mitigate data loss by leveraging alternative estimation sources when needed. However, the neural network is trained using the difference between UWB position and ground truth position to inflate the covariance, leading to the need for a large dataset for training. Therefore, the ground truth is always necessary for a new environment, which may sometimes be infeasible to collect.
An adjustment to an empirical variance model tailored for UWB range measurements based on time-of-flight (ToF) estimates has been introduced by \cite{cano2023robust}. This adjustment leverages the received first path power (FPP) to improve accuracy. Building on this foundation, they developed a robust M-estimation Kalman filter (M-RKF) to design an advanced UWB-based navigation system capable of effectively handling multipath (MP) outliers that often undermine measurement reliability.
Autoencoders are a prevalent tool in the field of novelty detection, largely due to their capacity to learn compact representations of data, as evidenced by the literature \cite{del2022novelty, zhao2023novelty}. In the context of UWB technology, they are employed for various purposes, including the assessment and correction of signal reliability. Variational autoencoders have been employed to assign anomaly scores to channels in ToF-based systems, exhibiting strong generalization abilities. These scores can be integrated into a Kalman filter-based position tracking system as in \cite{stahlke2022estimating}. However, this work employs the Channel Input Response (CIR) as inputs of the networks, which are extremely sensitive to noise and may represent a major limitation for real-time performances due to embedded hardware constraints. Furthermore, autoencoders have been employed to enhance the precision of ranging measurements, particularly in localization scenarios \cite{fontaine2020edge}.

\section{Methodology}
\label{sec:methodology}
UWB localization is based on ToA range measurements between a tag and multiple anchors. Unfortunately, these measurements can be affected by environmental elements such as metal obstacles and sources of electronic interferences, leading to errors in ToA computation. As a consequence, wrong ranges affect the entire positioning estimation task.

This paper aims to mitigate the issue of UWB-based localization by proposing an effective and general solution, particularly adapted and validated for small and noisy indoor environments where the limitation of ToA range estimation heavily hinders precise localization. The proposed solution exploits the potential of Deep Neural Networks for novelty pattern detection in range signals, combined with classic estimation and filtering techniques. The innovative localization framework is, therefore, composed of two main modules: a Novelty Detection Neural Network (NDNN) and an Extended Kalman Filter (EKF) algorithm.

The complete pipeline of the proposed localization system is shown in Figure \ref{fig:pipeline_schematic}. As a first preliminary step, an UWB ranges dataset is collected in the environment to establish the nominal conditions for the range measurements. The collected dataset is then used to train the NDNN, enabling the network to learn the feature pattern representing the nominal range values within the observed environment. Then, the output of the NDNN is compared with the input data by computing a novelty score, which represents the error between the predicted and observed range values. The novelty score is then remapped to a covariance and bias term for the correction step of the EKF, using simple activation functions. The robustness of model-based tracking algorithms such as the EKF is therefore mixed with NDNN to provide an adaptive system for multiple environments. The following sections provide a more detailed explanation of the working principle and structures of each framework component.

\subsection{Neural Network Architecture}
\label{subsec:NN_archi}

\begin{figure}[!t]
\centering
\includegraphics[trim={0 0.5cm 0 0.5cm}]{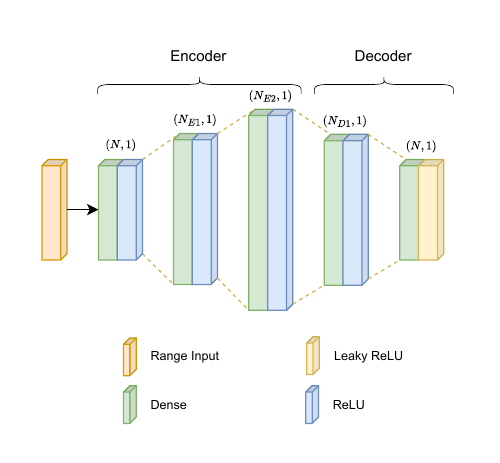}
\caption{The autoencoder architecture.}
\label{fig:autoenc_architecture}
\end{figure}
The neural network architecture used for the purpose of this work is a semi-supervised autoencoder. Autoencoders are a family of generative neural networks that learn to reconstruct the input by encoding it to a latent feature space, using an encoder, and decoding it back to the original input using a decoder. This latent space could have a higher or a lower dimension with respect to the input dimension. If the latent space has a higher dimension than the input, the autoencoder is called an overcomplete autoencoder. In contrast, if the latent space has a lower dimension than the input, it is called undercomplete autoencoder.
Formally, let $X \in \mathbb{R} ^ n$ represent the input data, where $n$ is the input's dimension. Let $f_{enc} \in \mathbb{R} ^ k$denote the encoding function transforming the input to the latent space, and $f_{dec} \in \mathbb{R} ^ n$ the decoding function transforming back the latent space to the original input. The complete formulation of the autoencoder is: 
\begin{equation}
    X_{recon} = f_{dec}(f_{enc}(X))
\end{equation}
In the overcomplete autoencoder $k>n$ while in the undercomplete autoencoder $k<n$. 

In this paper, we use an overcomplete autoencoder, meaning that the latent space has a higher dimension than the input space since it is harder for an undercomplete autoencoder to train properly when the input has a low dimension since it cannot learn the necessary features. The overcomplete architecture is shown in Figure \ref{fig:autoenc_architecture}. The network input comprises UWB ranges between the tag and the anchors, normalized between 0 and 1. First, the encoder transforms this input into a latent dimension space $k$. The encoder is composed of fully connected layers, where the number of neurons decreases progressively. Then, the decoder, which reverses the process, transforms the latent space back into an output with the same dimension as the input $n$. 
Formally, the number of neurons in the encoder follows this relationship: $N<N_{E1}<N_{E2}$ and in the decoder: $N_{E2}>N_{D1}>N$, where $N=n$ and $N_{E2}=k$. We use a ReLU activation function for all the layers except for the final one, which uses a Leaky ReLU function to avoid the issue of zero saturation.

\subsection{Data collection and novelty learning}
\label{subsec:data_collection}
In the majority of the previous works, UWB range errors are learned through a dataset collection and manual annotation \cite{karfakis2022uwb, poulose2021feature, nguyen2020convolutional}. With the proposed novelty method, a semi-supervised approach is adopted, avoiding the costly effort of manually labeling data.
Moreover, in this work, we demonstrate that it is possible to learn the complete range signals characterization in the environment, in nominal conditions, without the need to measure range values with a fine-grained fingerprint approach, as done in the latest study \cite{albertin2024case}. The novelty autoencoder has been trained using UWB range data collected from a few robot trajectories in the environment. This is a key enhanced feature of the approach that could also allow online learning and adaption of the neural network in future works.

\subsection{Novelty score computation}
\label{subsec:NN_score}
As explained in the previous paragraph, once the model is trained, the output of the network should match the input. A novelty score is used to quantify the severity of the difference between the input and the reconstructed output. The novelty score is defined as follows:
\begin{equation}
    e_{i} = |X_{recon,i} - X_{i}|
\end{equation}
where $X_{recon,i}$ is the reconstructed range of the $i^{th}$ anchor, while $X_{i}$ is the input range of the same $i^{th}$ anchor. The number of novelty scores obtained is equal to the number of anchors used. The idea relies on the observation that when the nominal conditions change, the neural network is no longer able to accurately reconstruct the input, causing a discrepancy between the input and the output. This mismatch leads to an increase in the novelty score, highlighting a novelty in that position. This discrepancy may arise from obstacles in the environment that interfere with the UWB signal.

\subsection{Adaptive robot localization with EKF}
\label{subsec:method_localize}
The overall system combines the novelty detection output previously discussed in Section \ref{subsec:NN_archi} and \ref{subsec:NN_score}, with a standard Extended Kalman Filter (EKF) approach for position tracking of a mobile robot. 
This research focuses on enhancing the adaptivity of UWB-based localization to any environmental disturbance caused by the static geometry and items in the scene, as well as any temporal changes in the environment. For static filter-based solutions, evolving environments and different signal disturbance conditions are challenging to handle. Achieving an optimal trade-off for the covariance values associated with sensor measurements is particularly difficult for mixed situations. On the other hand, for learning-based approaches, re-training on new datasets acquired from the new environmental condition would be necessary. Our method aims at dynamically reaching a sufficient level of flexibility and adaptation to previously unseen UWB range conditions by integrating the novelty information in the correction step of the EKF. Novelty scores computed for each UWB anchor by the autoencoder are exploited to estimate both a correction bias for the range measurements and the associated covariance values that regulate the expected uncertainty in data.

The state used in the EKF is straightforward: a constant velocity motion model is considered in a fixed absolute coordinate frame:

\begin{equation}
    \begin{aligned}
        x_{t+1} = x_{t} + v_x \cdot dt + \delta_x  \\
        y_{t+1} = x_{t} + v_x \cdot dt + \delta_y  \\
        v_{x,t+1} = v_{x,t} + \delta_{vx}      \\
        v_{y,t+1} = v_{y,t} + \delta_{vy}      \\
    \end{aligned}
\end{equation}

Where $x,~y$ and $v_x,~v_y$ are the positions and the velocities on the $X$ and $Y$ axis, respectively, and $\delta$ indicates the white noise used to model the uncertainty generated in the robot motion process due to inaccurate actuation errors or not modeled dynamics.

The measurement model of the UWB ranges is defined as the Euclidean distance $\| \epsilon \|_2$ between the estimated pose of the robot and each UWB anchor position in the fixed frame:

\begin{equation}
\begin{split}
    h_{rng}(\mu) & = \|(\mu - \gamma)\|_2 = \| \epsilon \|_2 = \sum_{k=1}^N{| \epsilon_k |^{1/2}} = \\ 
    & = \sqrt{(x_{r} - x_{uwb})^2 + (y_{r} - y_{uwb})^2 + (z_{r} - z_{uwb})^2}
\end{split}
\end{equation}
where $\mu$ and $\gamma$ are the robot pose and the UWB anchor pose, respectively.
The distance over the $z$ axis is constant for a wheeled robot moving in the $X-Y$ plane and having a tag mounted on it at a height different from the anchors, different from an aerial drone moving in $3D$ space.
In usual EKF implementation, the covariance associated with measurements is constant, and it is defined according to the a priori information available about sensor data reliability. The diagonal matrix $\mathbf{R}_t$ defined below contains the covariance of each UWB anchor's range, assuming no correlations between ranges.

\begin{equation}
\mathbf{R}_t =
  \begin{bmatrix}
    \sigma_{1,t}^2 & & \\
    & \ddots & \\
    & & \sigma_{a,t}^2
  \end{bmatrix} \\
\end{equation}

\begin{figure}[ht]
\centering
\includegraphics[width=10cm,trim={0 0.8cm 0 0.8cm}]{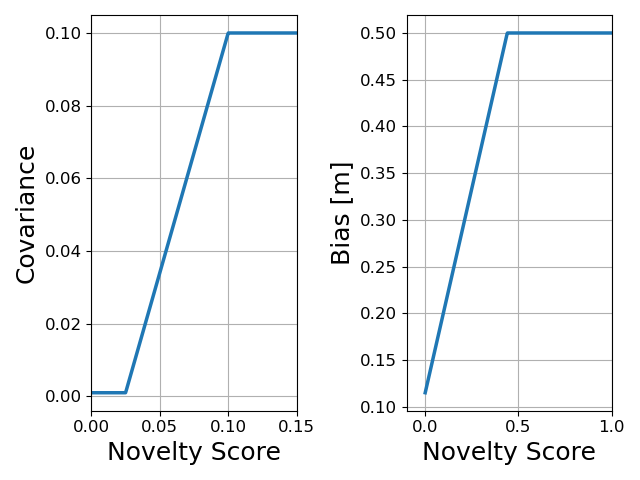}
\caption{Mapping functions used to transform the estimated novelty score into a bias (right) and a covariance (left) term for the UWB correction step of the Extended Kalman Filter. These functions are kept equal for all the experimental scenarios. The bias term is estimated from novelty scores using a mapping function able to refine the range value passed to the EKF.}
\label{fig:mapping_functionss}
\end{figure}

However, UWB ranges provide different range quality based on the geometrical and material properties of a specific environment. 
Hence, it is often hard to define single variance values that correctly depict the real confidence in ranges for the overall environments. The idea behind this work is, therefore, to dynamically set the covariance of the UWB measurements, mapping the estimated novelty score for each range to a covariance value. 
Figure \ref{fig:mapping_functionss} (left side) shows the simple mapping function used to translate the novelty score predicted by the neural autoencoder to a covariance value for the EKF correction step. The covariance computed from novelty scores can range from a low value of $0.001$, associated with highly confident data, and a maximum value of $0.1$ assigned to UWB ranges that present different values from nominal conditions and, consequently, higher uncertainty.

The estimated covariance informs the filter about the confidence to treat each UWB data, however, the correction step will be performed in any case with all range data, including not precise ranges affected by multiple reflections of the signal. Therefore, a bias error term can also be estimated from novelty scores using a different mapping function to refine the range value passed to the EKF. 
The linear transformation designed for this purpose is shown on the right side of Figure \ref{fig:mapping_functionss}. The shape of the mapping function is the same as the covariance one: a simple rectified linear mapping, but using different slope and saturation values, which have been chosen empirically based on average error discrepancy found in the training UWB data.
More in detail, it is derived by fitting a first-degree polynomial $y = mx + q$ on training data to model the relationship between the novelty score and the range error. The fitting process uses reconstructed ranges obtained from the Vicon system through inverse trilateration and raw UWB ranges measured. The terms $m$ and $q$ are computed for each anchor and then averaged.
The bias transformation coefficients have been found equal to $m = 0.885$ and $q = 0.115~m$, whilst the maximum error value for saturation is fixed to $0.5~m$. A non-null bias term is therefore assigned also for zero novelty scores. This is reasonable because real cluttered environments can already present a significant disturbance and reflection rate due to walls and obstacles, which are not embedded in the novelty score that is fitted on such nominal conditions. The saturation threshold is instead useful to cut out outliers with abnormal novelty score values.

These simple but general mapping functions represent a clear enhancement of our approach compared to previous works that attempt to estimate biases and covariances from data, relying on an external ground truth signal source \cite{che2022feature}. The experiments presented in Section \ref{sec:exp_results} demonstrate the competitive advantage of integrating such a simple error bias estimate obtained from the novelty score.

The correction method is therefore performed using the EKF equations:

\begin{align}
    \mathbf{K}_t = \mathbf{\Sigma}_t \mathbf{H}^T (\mathbf{H} \mathbf{\Sigma}_t \mathbf{H}^T + \mathbf{\hat{R}}_t)^{-1} \\
\mu_{t+1} = \mu_t + \mathbf{K}_t (z_t - \hat{\zeta}_t - \mathbf{H} \mu_t) \\
\mathbf{\Sigma}_{t+1} = (\mathbf{I} - \mathbf{K}_t \mathbf{H}) \mathbf{\Sigma}_t 
\end{align}

Here, the Kalman gain $\mathbf{K}_t$ is computed in the first equation, where $\mathbf{H}$ is the Jacobian of the measurement model over the state, and $\mathbf{\hat{R}}_t$ is the covariance matrix estimated from the novelty score of the UWB measurement $z_t$. 
The mean of EKF is therefore corrected using the innovation vector $\mathbf{K}_t (z_t - \hat{\zeta}_t - \mathbf{H} \mu_t)$ in the second equation, with the Kalman gain obtained with the dynamic covariance estimate and $\hat{\zeta}_t$ being the bias computed for the ranges $z_t$.

\section{Experiments and Results}
\label{sec:exp_results}

This section provides a detailed description of the experiments conducted and the laboratory settings adopted, followed by a thorough presentation of the system results. Experiments and results are reported and discussed starting from the NDNN application on UWB signal. Next, findings on robot positioning are presented based on tests in various environmental scenarios. Finally, the section explores an ablation study examining the impact of key factors on localization performance, including the number of UWB anchors and the individual benefits of dynamic covariance and bias correction for the filter.

\subsection{Experimental Setup}
\label{subsec:exp_setup}
\begin{figure}
    \centering
    \includegraphics[width=10cm]{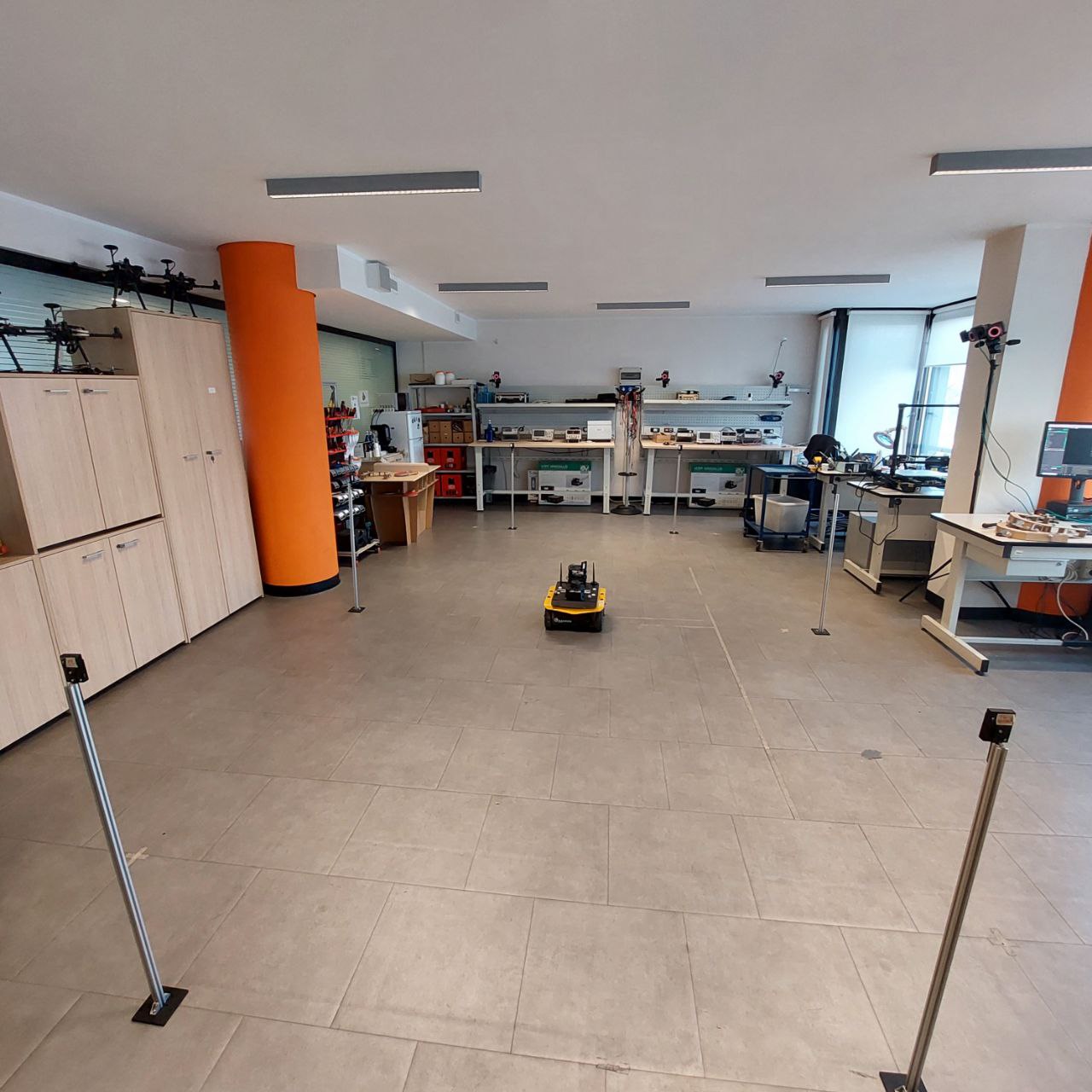}
    \caption{The laboratory equipped with a Vicon tracking system and the Jackal rover used for the experiments. Obstacles are placed in different positions around the area to create the testing scenarios.}
    \label{fig:lab}
\end{figure}

\begin{figure}[t]
\centering
\includegraphics[width=10cm,trim={0 1cm 0 1cm}]{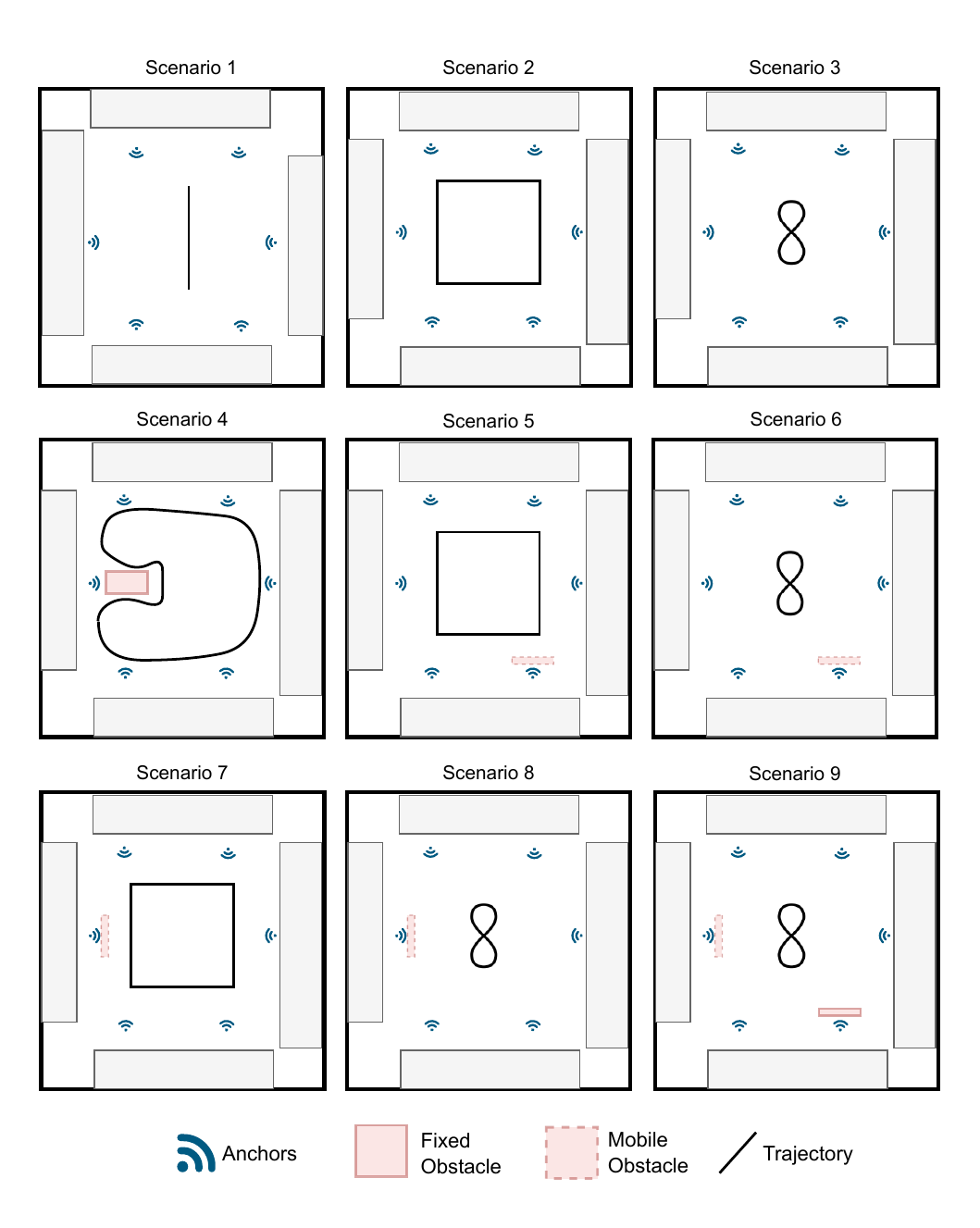}
\caption{Testing scenarios designed for this paper. The first three scenarios represent distinct trajectories conducted under LoS conditions.}
\label{fig:Experiments}
\end{figure}

The tests were conducted in the laboratory room environment at the PIC4SeR center. The robot was moved within a rectangular area with an approximate size of $6~m$ by $3~m$, surrounded by UWB anchors. The robot used in the experiments was a Jackal from Clearpath Robotics. The maximum velocity reached by the robot is set to $1~m/s$. The ground truth for the framework was obtained using a Vicon Tracker System, recording the complete trajectories of the robot. The office room of tests is shown in Figure \ref{fig:lab}. Small indoor spaces disrupt UWB signals due to reflections and distortions from walls and office furniture like chairs and tables. Metal parts are the main cause of these issues, affecting communication between the tag and the anchor. We used 6 UWB anchors in a hexagonal configuration (see Figure \ref{fig:Experiments}) for the main experiments presented in Section \ref{subsec:localization_test}. In the ablation study, the number of anchors has been then varied from 6 (in a hexagonal configuration) to 4 (in a square configuration). 

The experiments phase considers nine different scenarios shown in Figure \ref{fig:Experiments}. In this phase, the trajectories are designed to better capture the effects of UWB signal corruption in the environment, evaluating the performance of the adaptive novelty-based filtering technique. The first three scenarios involve nominal environmental conditions, differing from the NDNN training phase only in the trajectories performed. In scenario number 4, the trajectory is random, and the environment has a fixed obstacle on one side of the rectangle. In scenarios 5 and 6, as well as 7 and 8, the same trajectories are performed while the NLoS conditions are created by occluding a different anchor during the test. Finally, scenario 9 shows the most challenging condition, placing a fixed obstacle in front of an anchor and a dynamic obstacle on another anchor during the test execution. Hence, scenarios 5, 6, 7, 8, and 9 include various trajectories with dynamically changing NLoS conditions, a challenging aspect of real-world UWB applications that is rarely addressed in the existing literature. In these scenarios, NLoS conditions are introduced during testing by placing a metal plate in front of an anchor. This setup simulates potential real-world changes requiring traditional filtering methods to be re-tuned and learning-based methods to be re-trained. Differently, the obstacle of scenario 4 is a metal cabinet placed on the side of the map, creating a positional NLoS condition depending on the position of the robot. On average, the trajectories collected last 50 s, except for scenario 4, which lasts 80 seconds. Overall, the experiment design illustrated aims at demonstrating that learning a characterization of the environment represents a substantial advantage in UWB-based positioning systems, and adaptive localization solutions can outperform static methods in such realistic cases never explored in detail before. A table (Table \ref{tab1}) is also shown to summarize the characteristics of each scenario, providing also the averages of the Novelty scores depending on NLoS condition.
 
\input{Table_scenarios}

\subsection{Novelty Detection on UWB signal}
\label{subsec:NN_test}

In this section, a graphical representation of the results obtained on UWB signals using the NDNN presented in Section \ref{subsec:NN_archi} is given. The hyperparameters of the Neural Network used are chosen by conducting a grid search with the following values: $E1,D1$ Neuron - $[15,20]$, $E2$ Neuron - $[20,30,40]$, Batch Size - $[16,32,64]$, Learning Rate - $[0.001,0.01]$. The best overcomplete structure obtained after the optimization has: $E1,D1 = 15$ neurons, $E2 = 30$ neurons, batch size  $=~32$, and learning rate $=~0.001$. The neurons of the first and last layer are equal to the number of anchors used. The training is performed with only 10 epochs. The average inference time with this architecture is $25~ms$ using an Intel Core i7-7500U CPU.

The training of the NDNN is conducted using data collected by moving the robot along random trajectories within the rectangle defined by the anchors' position. The success of the training using data from random robot trajectories represents a significant advantage over the fingerprint approach presented in \cite{albertin2024case}, enabling a faster collection of the training dataset. Rather than taking multiple measurements for each single position, this approach collects a single measurement across multiple positions. The dataset could be considered consistent once the robot has covered all the rectangle positions. 

\begin{figure}[ht]
\centering
\includegraphics[width=10cm,trim={0 0.5cm 0 0.5cm}]{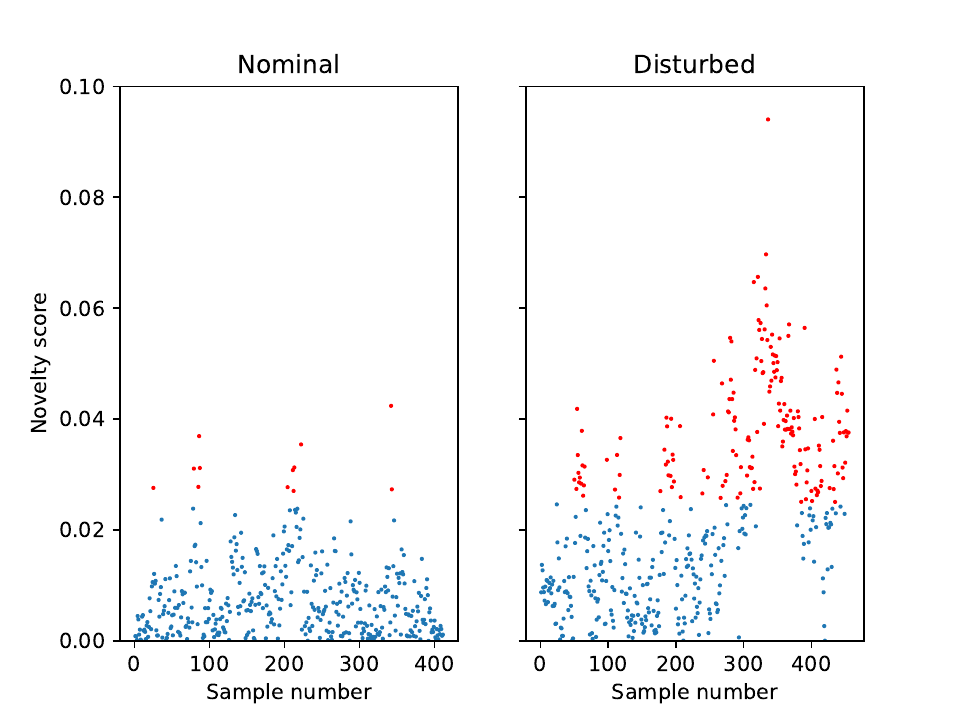}
\caption{The outputs of the novelty network. The novelty score represents the error between the network's predictions and the raw ranges fed into the network. In nominal scenario 3 (left graph), the errors are lower in all the sample points than in the disturbed scenario 9 (right graph).}
\label{fig:nomvsdist}
\end{figure}

An example of the novelty score obtained by receiving a UWB signal from an anchor is shown in Figure \ref{fig:nomvsdist}, illustrating the difference between nominal and disturbed conditions. The plots reported are related to the outputs of an anchor in scenarios 3 (left) and 9 (right). The novelty score remains low under nominal conditions in scenario 3, increasing when the disturbance is introduced in scenario 9. Some outliers appear in the plot due to random errors of the UWB antennas we use for testing, producing sparse wrong range measurements. In scenario 9, the effectiveness of our NDNN solution is shown, highlighting the beginning NLoS condition due to an occluded anchor (till the $200^{th}$ sample), followed by a further degradation when a metal plate is placed in front of another anchor. This additional obstruction further reduces the quality of the measurements, leading to an increase in the error between the network's input and output. The plot also shows that the values chosen to design the EKF covariance and bias activation map shown in Figure \ref{fig:mapping_functionss} fit well the resulting data. Values below $0.025$ (marked by blue dots) are appropriate for identifying conditions close to the nominal ones. Using these values, scenario 3 has most of the points in LoS conditions as it is in reality. In contrast, scenario 9 has a lot of points identified as NLoS due the the double anchors occlusion. A maximum threshold of $0.1$ is used to saturate the covariance, reflecting the increased uncertainty caused by disturbances in the UWB measurements. This kind of correction is done for each anchor used for localization. Moreover, also a bias activation map has been designed to map the predicted novelty score to a direct range correction. When the NDNN detects a nominal Line-of-Sight (LoS) condition (near-zero novelty score), a correction of about $11cm$ is applied to the range measurements. In contrast, when a high Non-Line-of-Sight (NLoS) condition is identified by the NDNN, a $50cm$ correction is applied. This bias adjustment accounts for the noisy nature of small indoor environments, where distortions caused by walls and office furniture affect measurements even under nominal conditions. This approach effectively mitigates the common issues affecting UWB antennas in such environments.

\begin{figure*}[t]
\centering
\includegraphics[width=15cm,trim={0 0.5cm 0 0.5cm}]{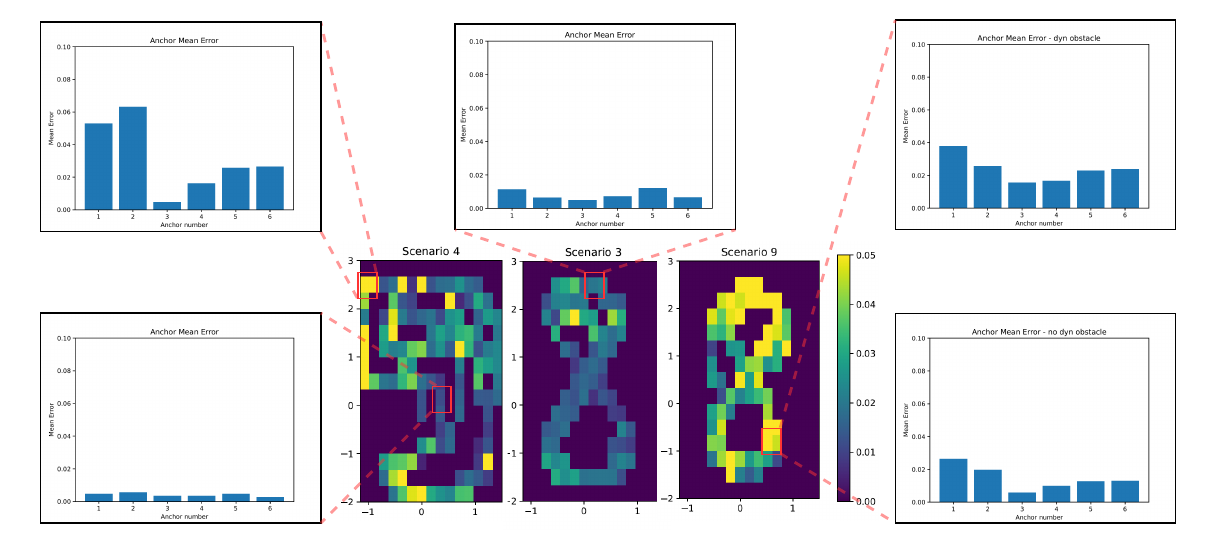}
\caption{Novelty network's scores heatmaps for scenarios 3, 4, and 6. The more yellow zones show big differences between measured ranges and nominal ones used for training.}
\label{fig:ndoutput}
\end{figure*}

In Figure \ref{fig:ndoutput}, three representative cases are shown to provide a clearer understanding of how the proposed novelty detection framework works. The heatmaps show three different testing scenarios, namely a nominal condition (scenario 3) and two disturbed conditions (scenarios 4 and 9). The squares of the grid map are obtained by taking several positional intervals by subdividing the whole map into 16 squares - long 0.18m in $X$ and high 0.33m in $Y$ - and averaging all the range measurements within this interval in $X$ and $Y$. In scenario 3, at the center of the figure, the novelty scores remain low in most of the trajectories performed due to consistently maintained LoS conditions. This scenario is always in nominal conditions, resulting in a heatmap predominantly with low values of the novelty score in blue and some higher values in yellow, appearing in the middle of the map. These yellow regions represent novelty points, i.e., outliers with respect to the nominal conditions, which are not removed in order to highlight the framework's ability to handle such anomalies. These outliers could occur in real-world situations due to communication issues between the tag and the anchors, and multi-path reflections caused by the environment. In such cases, even under LoS conditions, the UWB ranges used for the estimation of the position present an error that should be mitigated by correcting it or raising the covariance value associated with the measurement. Moreover, the histogram plots show a detailed anchor-specific view of the average error in representative points of the gridmap. In scenario 4, on the left of the figure, it is possible to appreciate how different map regions can present totally different novelty scores and, therefore, errors in the UWB ranges. These findings strongly reinforce the founding hypothesis of the proposed method: learning a precise characterization of the environment is crucial for UWB localization. Indeed, regions in the center and right side of map 4 show low novelty scores, where LoS conditions with all anchors are preserved. However, in other areas, specifically in the up and bottom left parts of the trajectory, the novelty scores are higher due to a metal cabinet that creates UWB noise and NLoS conditions. As explained, this scenario demonstrates the ability of the proposed method to identify NLoS conditions position-wise in a map using the novelty score. Regions with high range errors are predicted thanks to the discrepancy from training to testing conditions in the received range signal.

\begin{figure}[ht]
\centering
\includegraphics[width=12cm]{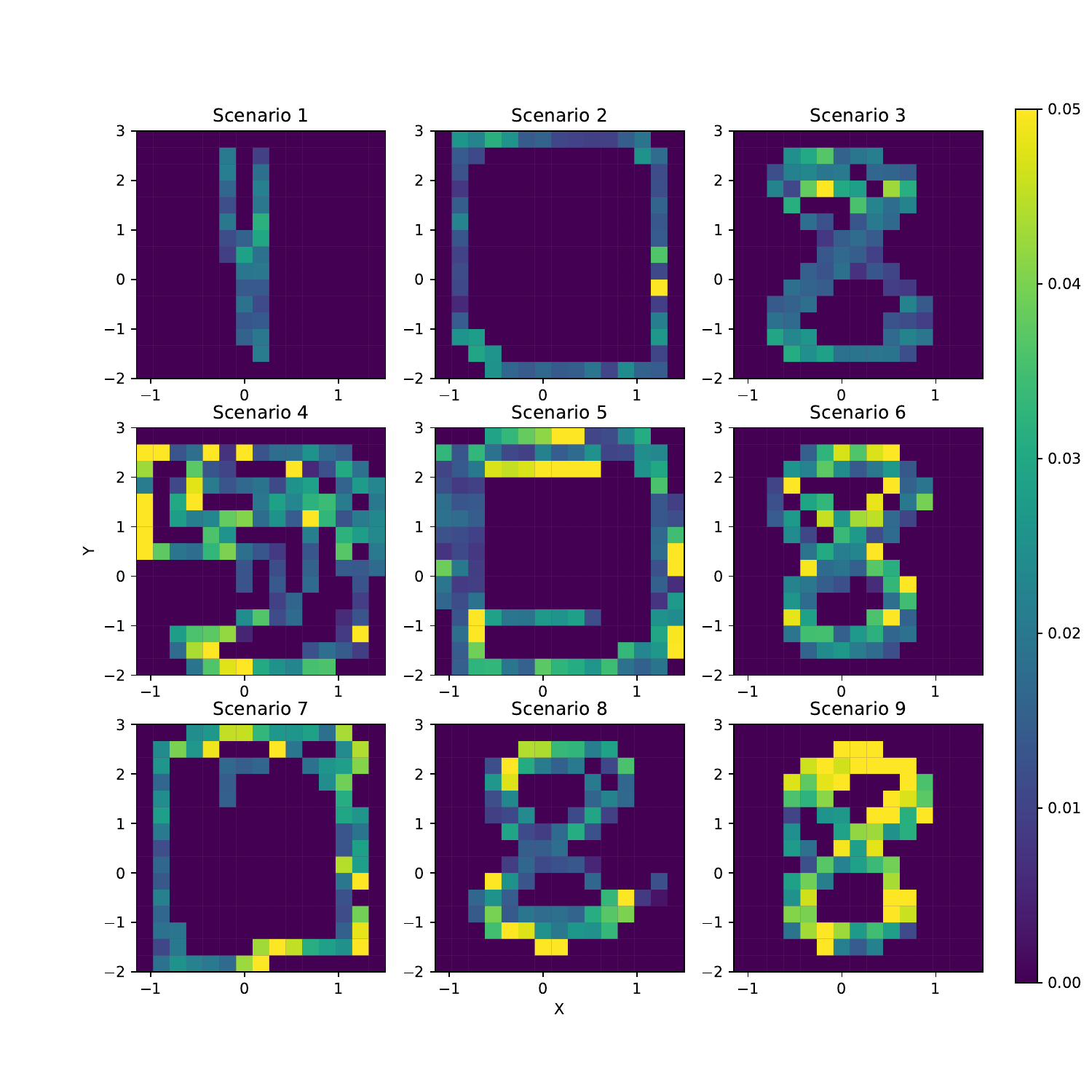}
\caption{Heatmaps for each scenario related to the novelty score}
\label{fig:allheat}
\end{figure}

Scenario 9 represents the most challenging condition, with two anchors occluded during the test, resulting in NLoS conditions for most positions in the trajectory. The detailed view of the anchor's behavior reports the novelty score histograms of the same point in the map at different time instants. The proposed NDNN demonstrates the ability to identify different NLoS levels over time, showing a dynamic increase in the novelty score due to a plate placed in front of an anchor during the test execution.
In this complex case, however, the NDNN struggles to reconstruct precisely all the anchors' ranges individually, given that the NLoS conditions are highly present in the scene. This is the reason why the error is generally higher in all the anchors rather than only in the affected ones. Nonetheless, the main goal of the novelty framework is to inform the filter about general NLoS conditions in a certain zone of the map to mitigate the propagation of the error in the pose estimation phase. 

This analysis is performed for each testing scenario, and a summary is shown in Figure \ref{fig:allheat}. As shown from the plots, the first three scenarios are in LoS conditions, obtaining a low novelty score represented by blue regions in the grid map. Scenarios 5 and 7 show a similar behavior since the test has been performed similarly; a different anchor is occluded during the test, showing a consistent NDNN output for both experiments. Comparable results are obtained for scenarios 6 and 8 but with different trajectories. From this overall graphical comparison, the differences between the occlusion of one anchor and two anchors clearly emerge. For example, comparing scenarios 6 or 8 with scenario 9, it is possible to notice a higher novelty score in the majority of the points of scene 9.

\subsection{Robot Localization}
\label{subsec:localization_test}

\input{Table_uwb_bias}

The overall localization system presented in \ref{subsec:method_localize} has been tested on nine different experimental scenarios. Moreover, in the ablation study proposed at the end of the section, these nine scenes are combined with three different anchor configurations, resulting in 27 total testing conditions adopted to demonstrate the advantages of the proposed solution. The nominal scenario is the office room shown in Figure \ref{fig:lab}. 
The Nov-EKF proposed solution has been compared with the standard EKF using static values for the covariance associated with the UWB ranges in the correction step of the filter. A value of $0.01$ is used for the UWB covariance of the static EKF, resulting as the optimal trade-off value over all the scenarios.

\begin{figure*}[ht]
\centering
\includegraphics[width=15cm,trim={0 2.5cm 0 2.2cm}]{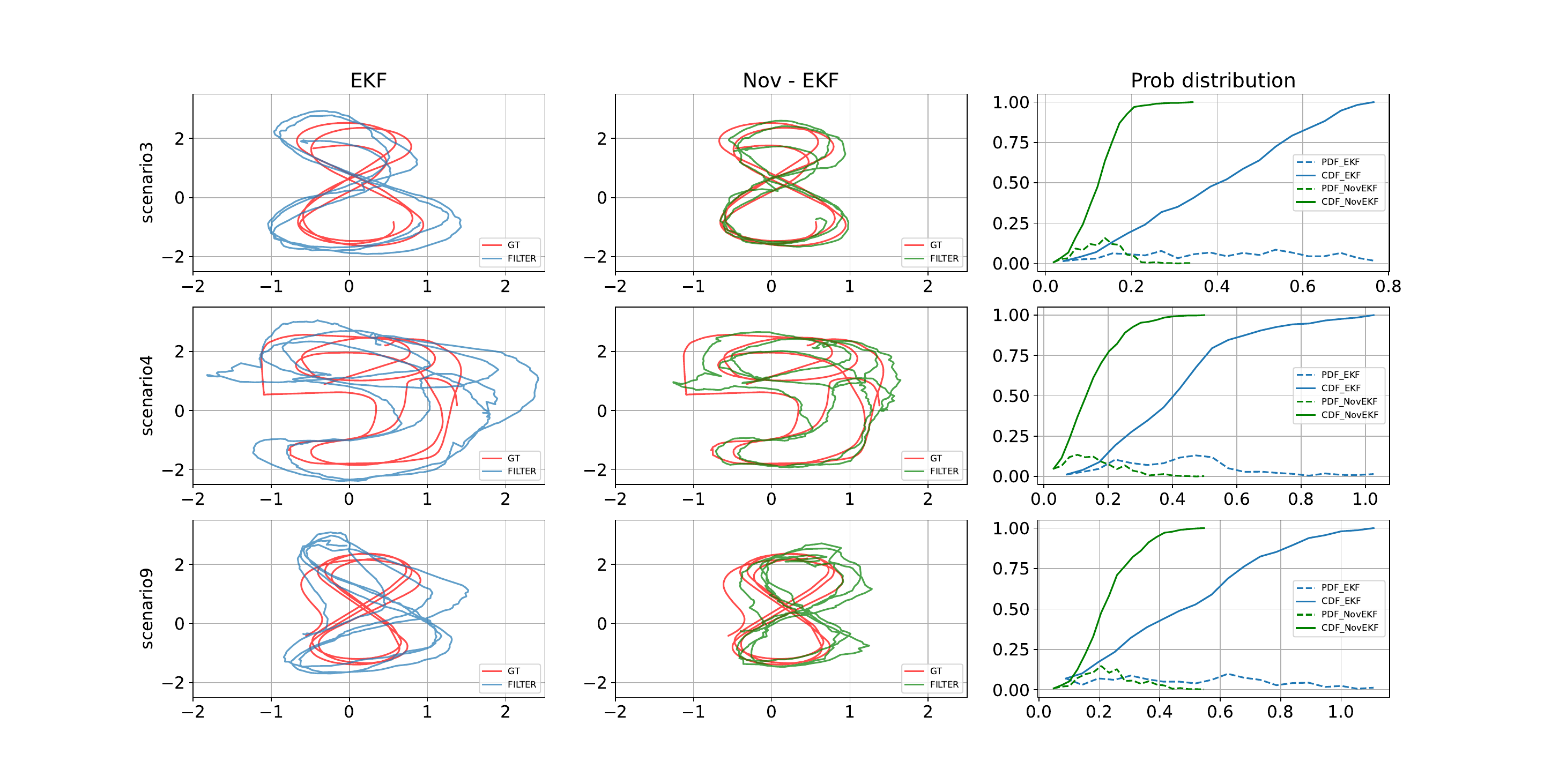}
\caption{Trajectories reconstruction of scenarios 3, 4 and 9. Standard EKF trajectory reconstruction on the left, our Nov-EKF framework reconstruction in the middle, and CDF/PDF between the two approaches on the right.}
\label{fig:aggregate_res}
\end{figure*}

\begin{figure*}[ht]
\centering
\includegraphics[width=16cm,trim={0 1.5cm 0 1.5cm}]{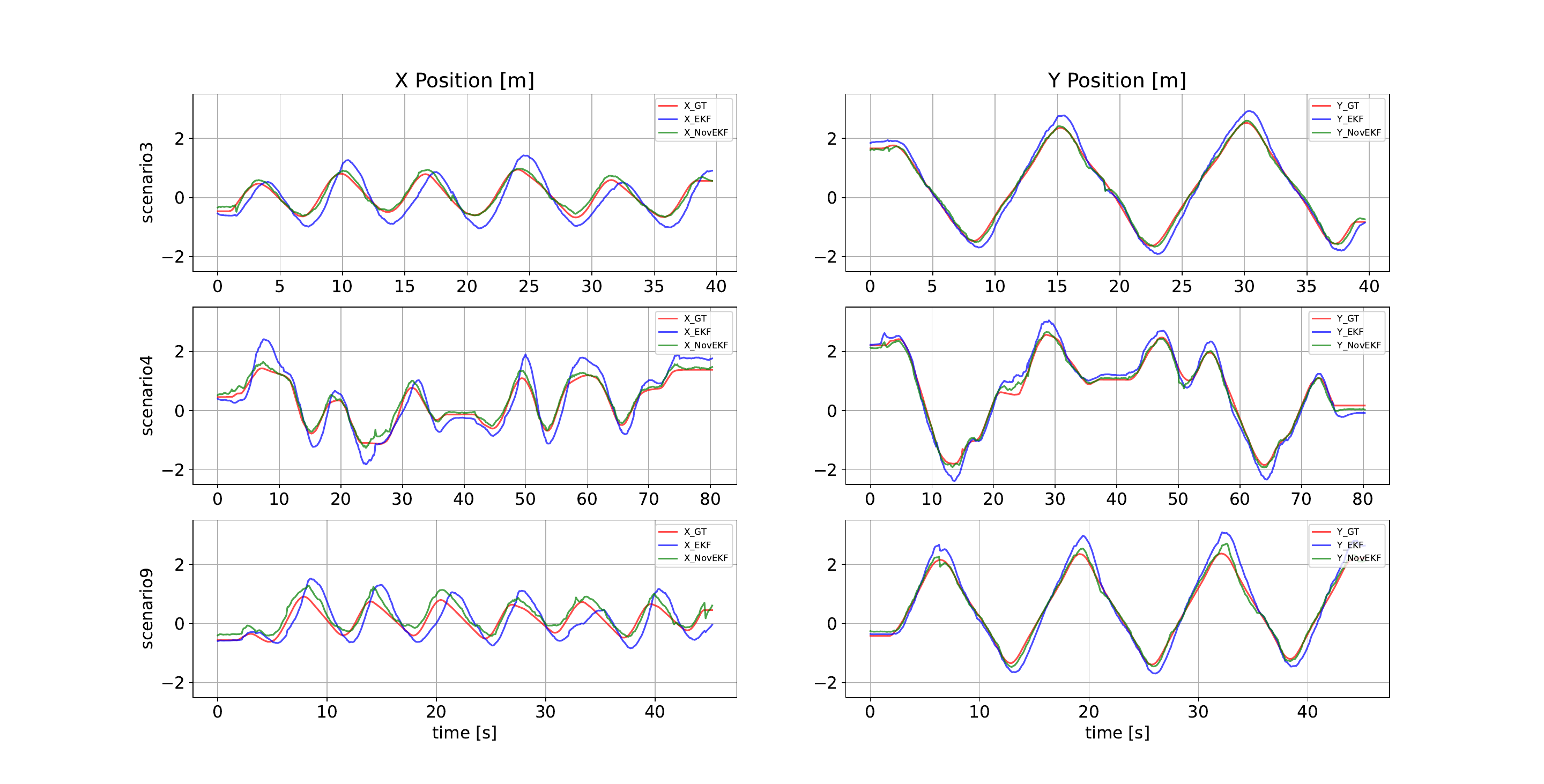}
\caption{X and Y positions comparison between ground truth, EKF and our framework.}
\label{fig:xy_error}
\end{figure*}

First, a quantitative summary of the positioning results obtained is reported in Table \ref{tab2}. In this table, Root Mean Squared Error (RMSE) and  Mean Absolute Error (MAE) are computed for each scenario using the ground truth trajectory of the robot and comparing the classic EKF with our novelty-based approach. As can be noticed from the values in the table, Nov-EKF drastically reduces errors compared to the baseline in all the scenarios, considering both soft NLoS and hard NLoS conditions. For both single and average testing results, the bottom line of the table, RMSE, and MAE metrics are often more than halved, showing remarkable improvements in accuracy and robustness and also in challenging real settings.
For a better visualization and interpretation of the performance obtained, the same three scenarios 3, 4, and 9 analyzed in the previous sections are considered for localization purposes. Figure \ref{fig:aggregate_res} shows the trajectories obtained with the whole Nov-EKF system and the static EKF, both compared to the ground truth collected with the Vicon cameras. 
From a general point of view, the trajectories obtained with the adaptive Nov-EKF are more accurate and closer to the ground truth than the trajectories obtained with the static EKF alone across all analyzed cases. The error increases as the NLoS level becomes stronger in the scene. This is evident when comparing scenarios 3 and 9, which have the same trajectory shape but drastically different results due to the presence of a double anchor occlusion in scene 9. In scenario 4, the error is focused on the top left region of the map, where the NLoS condition caused by the static obstacle is heavier, confirming the score shown in the heatmaps in Figure \ref{fig:allheat}. As an additional visual representation of the positioning error, the Probability Density Function (PDF) and the Cumulative Density Function (CDF) are shown on the right part of the figure for each scenario, comparing the static EKF with the adaptive Nov-EKF. The two distributions are computed using the error obtained by subtracting the correction output from the ground truth. A greater CDF means lower errors and, hence, better performance. Similarly, a PDF peak shifted towards the origin of the axis means a more frequent lower error with the ground truth. As shown in the plots, all the CDFs obtained with our framework Nov-EKF are higher than those of the EKF, and all the PDFs have peaks on the left side of the plot, highlighting a better behavior of our approach compared to the EKF alone. 
Finally, in Figure \ref{fig:xy_error}, a detailed signal plot of $X$ and $Y$ position predictions is shown for the same representative scenarios, highlighting the errors between the ground truth and the predictions in $X$ and $Y$ coordinates. The green line represents our proposed solution Nov-EKF, the blue one is the static EKF estimation, and the red one is the ground truth. These detailed graphical trends remark on the boost in accuracy provided by the adaptive novelty-based filter, reaching very similar to the ground truth and minimizing the overshoot in turning points compared to the static EKF. This evident improvement is mainly related to the adaptive covariance and bias assigned in the correction step to UWB range data, which represent the most impacting information for the filter.

\subsection{Ablation study}
\label{subsec:ablation}

In this section, we propose an ablation study to provide an in-depth analysis of the performance of the proposed Nov-EKF system. The goal of this study is to evaluate the influence of relevant parameters and settings on the overall capabilities of the solution. To do so, we conducted 27 experiments involving 4, 5, and 6 anchor configurations in all nine scenarios. The trajectories and obstacle positions remain unchanged also for the UWB configurations with 4 and 5 anchors, respectively. The side anchors of Figure \ref{fig:Experiments} are removed, transforming the initial hexagonal shape anchors configuration to a square shape with four anchors. 
With this ablation study, we focus on analyzing the effect and contribution of two key elements:
\begin{itemize}
    \item The number of anchors: we aim to evaluate the system by varying the number of anchors placed in the environment to simulate also worst-case scenarios. When using only four anchors, frequent NLoS conditions may occur, leading to unreliable trilateration and localization issues. 
    \item Bias - Covariance: as described in the previous chapters, the adaptive Nov-EKF pipeline dynamically estimates both a bias and a covariance value for UWB range measurement for the EKF correction step. In this second ablation, we try the adaptive EKF with and without the estimation of the bias term from the novelty score, and we demonstrate the accuracy boost it provides. The bias activation mapping function is indeed more difficult to define from novelty score only compared to the covariance, and it may require access to ground truth data for a precise design of the polynomial coefficients.
\end{itemize}

\begin{figure}[t]
\centering
\includegraphics[width=15cm,trim={0 1.5cm 0 1.5cm}]{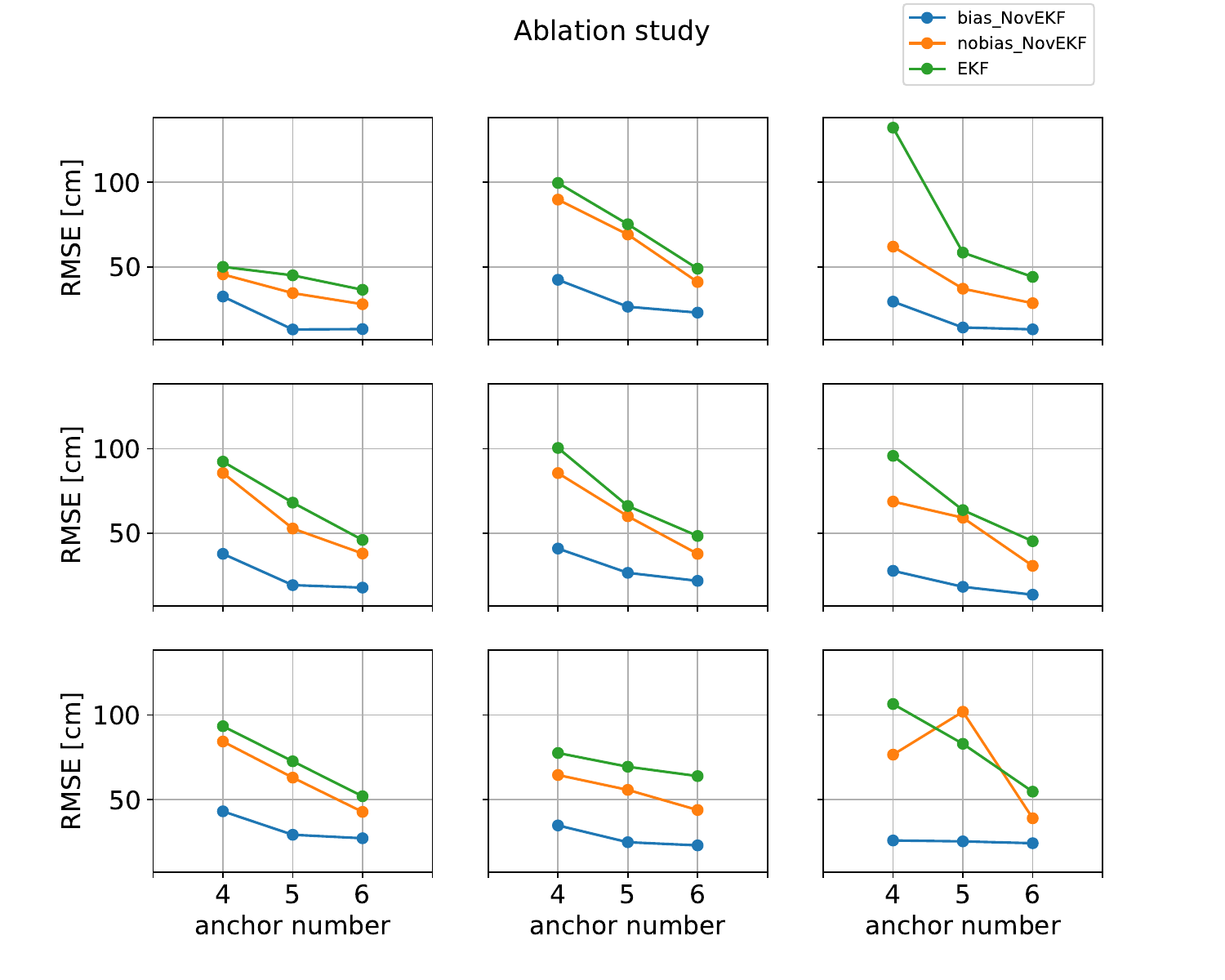}
\caption{RMSE metrics of the ablation study, comparing bias and nobias correction and analyzing the dependency with the number of anchors.}
\label{fig:ablation_bias}
\end{figure}

\noindent In Figure \ref{fig:ablation_bias}, the results of the ablation study are shown in a compact graphical representation where each subplot corresponds to one of the 9 scenarios. The points in the plots are the average RMSE values computed with respect to the ground truth for each of the three compared algorithms: static EKF (green), adaptive Nov-EKF without bias term estimation (orange), Nov-EKF with both covariance and bias terms estimations (blue).
Overall, it is possible to notice that the adaptive Nov-EKF achieves improved results in each configuration, also with covariance estimation only. On the other hand, a significant boost is provided by the introduction of the bias correction term in all the scenarios, also demonstrating a more robust behavior with a reduced number of anchors. The introduction of an adaptive correction term on the range is motivated by the fact that the novelty score represents itself a rescaled estimate of a range error, and it can provide such information. Nonetheless, it is fitted on nominal conditions, which in real cluttered environments can already present a significant disturbance and reflection rate due to walls and surfaces. Hence, also for low novelty scores, the introduction of a correction term on the ranges is absolutely beneficial for the filter.

The reduction of the number of anchors from 6 to 5 and 4 significantly affects the resulting performance of the filter. The static EKF and no-bias Nov-EKF especially lose prediction accuracy due to the missing redundancy in trilateration that can be exploited with 6 anchors. However, the complete Nov-EKF system shows more stable results on this side, with no substantial changes using 5 or 6 anchors in all the scenes. 
It can be generally said that a sufficient number of anchors is required to achieve reliable results in NLoS conditions. Among all the scenarios, only scenes 1 and 8 present a lower dependence on the number of anchors, meaning that in such cases, a minimum number of anchors in LoS is always guaranteed. 
Unexpected results are obtained in scenario 9, where a strong NLoS condition is caused after the occlusion of two anchors. The complete Nov-EKF system shows no substantial changes in this case, also with four anchors only. Hence, the system is able to significantly reduce the localization error with respect to classic EKF also with just two anchors in LoS. A different behavior is obtained when using 5 anchors without bias correction, causing the worst results in this scenario. This problem could be related to a too-generalized covariance estimation in the presence of high overall disturbances in the range signals, with the system not being able to discern between clean and dirty ranges on the specific anchors.

\section{Conclusions and Future Works}
This paper aims to advance robot localization using UWB technology by introducing the adaptive EKF-novelty technique. Our work demonstrates how to handle unreliable signals coming from noisy UWB measurements in indoor cluttered spaces presenting NLoS conditions, reaching an average improvement more significant than $25~cm$ of absolute positioning error.

This technique enables the EKF to discern when incoming measurements lack sufficient reliability using the covariance and bias term derived from novelty score estimation. The conducted experimentation shows the substantial advantages in terms of localization error provided by our method compared to the standard EKF baseline. Through the ablation study, we also evaluated how the number of UWB anchors and the bias term estimation can influence the framework's robustness, either enhancing or worsening their reliability.

In future work, online training on the novelty autoencoder can be introduced for further adaptation to environmental changes. Moreover, the EKF can be replaced with a factor graph or with a second neural network receiving sensor data with the associated novelty score to overcome the limitation of EKF regarding time horizon prediction. Additionally, we aim to experiment with sensor fusion and generative models to further enhance the prediction of the novelty score.

\section*{Acknowledgments}
\noindent This work has been developed with the contribution of the Politecnico di Torino Interdepartmental Centre for Service Robotics (PIC4SeR\footnote{\url{https://pic4ser.polito.it/}}).

This publication is part of the project NODES which has received funding from the MUR – M4C2 1.5 of PNRR funded by the European Union - NextGenerationEU (Grant agreement no. ECS00000036).

This publication is part of the project PNRR-NGEU which has received funding from the MUR – DM 351/2022

\bibliographystyle{unsrt}  
\bibliography{biblio}

\end{document}

%% file: Table_scenarios.tex
\begin{table}
\begin{center}
\caption{Scenarios description with the related NLOS conditions}
\label{tab1}
\begin{tabular}{ c  c  c  c }
\toprule
\textbf{Scenario} & \textbf{Trajectory} & \textbf{NLOS} & \textbf{Average ND}  \\
& &  \textbf{Condition} & \textbf{Scores}\\
\midrule
1& Line & LOW & 0.018\\
\midrule
2& Rectangle & LOW & 0.017\\ 
\midrule
3& Infinite & LOW & 0.019\\
\midrule 
4& Random & HIGH & 0.032\\
\midrule 
5& Rectangle & MEDIUM & 0.025\\
\midrule 
6& Infinite & MEDIUM & 0.028\\
\midrule 
7& Rectangle & MEDIUM & 0.026\\
\midrule 
8& Infinite & MEDIUM & 0.023\\
\midrule 
9& Infinite & HIGH & 0.042\\
\bottomrule 
\end{tabular}
\end{center}
\end{table}

%% file: Table_uwb_bias.tex
\begin{table*}[]
\begin{center}
\caption{RMSE and MAE values obtained for all the scenarios with both EKF and the proposed Nov-EKF.}
\label{tab2}
\begin{tabular}{cccccccc}
\toprule
\textbf{Scenario}  & \textbf{Algorithm} & \textbf{\begin{tabular}[c]{@{}c@{}}RMSE\_x\\ {[}cm{]}\end{tabular}} & \textbf{\begin{tabular}[c]{@{}c@{}}RMSE\_y\\ {[}cm{]}\end{tabular}} & \textbf{\begin{tabular}[c]{@{}c@{}}RMSE\_tot\\ {[}cm{]}\end{tabular}} & \textbf{\begin{tabular}[c]{@{}c@{}}MAE\_x\\ {[}cm{]}\end{tabular}} & \textbf{\begin{tabular}[c]{@{}c@{}}MAE\_y\\ {[}cm{]}\end{tabular}} & \textbf{\begin{tabular}[c]{@{}c@{}}MAE\_tot\\ {[}cm{]}\end{tabular}} \\ \toprule
Training           & EKF             & 38.5                                                                & 24.8                                                                & 45.8                                                                  & 32.7                                                               & 20.5                                                               & 41.6                                                                 \\ 
  & \textbf{Nov-EKF}       & \textbf{9.6}                                                        & \textbf{8.2}                                                        & \textbf{12.6}                                                         & \textbf{7.8}                                                       & \textbf{6.6}                                                       & \textbf{11.0}                                                        \\ \midrule
1                  & EKF             & 20.1                                                                & 30.5                                                                & 36.6                                                                  & 16.3                                                               & 26.6                                                               & 32.7                                                                 \\ 
        & \textbf{Nov-EKF}       & \textbf{6.4}                                                        & \textbf{11.7}                                                       & \textbf{13.3}                                                         & \textbf{5.4}                                                       & \textbf{8.9}                                                       & \textbf{11.4}                                                        \\ \midrule
2                  & EKF             & 42.3                                                                & 24.7                                                                & 49.0                                                                  & 31.3                                                               & 21.0                                                               & 42.8                                                                 \\ 
        & \textbf{Nov-EKF}       & \textbf{20.8}                                                       & \textbf{10.0}                                                       & \textbf{23.1}                                                         & \textbf{16.5}                                                      & \textbf{8.4}                                                       & \textbf{19.9}                                                        \\ \midrule
3                  & EKF             & 36.7                                                                & 24.5                                                                & 44.1                                                                  & 31.8                                                               & 19.7                                                               & 39.7                                                                 \\ 
       & \textbf{Nov-EKF}       & \textbf{10.0}                                                       & \textbf{8.6}                                                        & \textbf{13.2}                                                         & \textbf{8.2}                                                       & \textbf{7.1}                                                       & \textbf{12.2}                                                        \\ \midrule
4                  & EKF             & 38.0                                                                & 26.1                                                                & 46.1                                                                  & 31.6                                                               & 21.3                                                               & 41.2                                                                 \\ 
         & \textbf{Nov-EKF}       & \textbf{13.4}                                                       & \textbf{12.0}                                                       & \textbf{18.0}                                                         & \textbf{10.8}                                                      & \textbf{8.5}                                                       & \textbf{14.8}                                                        \\ \midrule
5                  & EKF             & 42.5                                                                & 23.3                                                                & 48.5                                                                  & 32.2                                                               & 18.8                                                               & 41.8                                                                 \\ 
        & \textbf{Nov-EKF}       & \textbf{19.3}                                                       & \textbf{10.4}                                                       & \textbf{22.0}                                                         & \textbf{15.2}                                                      & \textbf{8.9}                                                       & \textbf{18.9}                                                        \\ \midrule
6                  & EKF             & 37.2                                                                & 25.9                                                                & 45.3                                                                  & 31.1                                                               & 19.8                                                               & 39.7                                                                 \\ 
        & \textbf{Nov-EKF}       & \textbf{9.4}                                                        & \textbf{10.1}                                                       & \textbf{13.8}                                                         & \textbf{7.3}                                                       & \textbf{8.2}                                                       & \textbf{12.4}                                                        \\ \midrule
7                  & EKF             & 46.0                                                                & 24.3                                                                & 52.0                                                                  & 34.9                                                               & 20.1                                                               & 44.6                                                                 \\ 
        & \textbf{Nov-EKF}       & \textbf{25.6}                                                       & \textbf{9.3}                                                        & \textbf{27.2}                                                         & \textbf{20.8}                                                      & \textbf{7.6}                                                       & \textbf{23.3}                                                        \\ \midrule
8                  & EKF             & 59.1                                                                & 24.2                                                                & 63.9                                                                  & 43.7                                                               & 20.2                                                               & 51.4                                                                 \\ 
         & \textbf{Nov-EKF}       & \textbf{20.7}                                                       & \textbf{10.0}                                                       & \textbf{23.0}                                                         & \textbf{15.7}                                                      & \textbf{7.8}                                                       & \textbf{18.7}                                                        \\ \midrule
9                  & EKF             & 44.4                                                                & 32.0                                                                & 54.7                                                                  & 37.0                                                               & 25.2                                                               & 48.0                                                                 \\ 
        & \textbf{Nov-EKF}       & \textbf{21.8}                                                       & \textbf{10.6}                                                       & \textbf{24.3}                                                         & \textbf{19.6}                                                      & \textbf{8.6}                                                       & \textbf{22.5}                                                        \\ \midrule
\textbf{Avg\_Test}          & EKF             & 40.7                                                                & 26.2                                                                & 48.9                                                                  & 32.2                                                               & 21.4                                                               & 42.5                                                                 \\ 
& \textbf{Nov-EKF}       & \textbf{16.4}                                                       & \textbf{10.3}                                                       & \textbf{19.8}                                                         & \textbf{13.3}                                                      & \textbf{8.2}                                                       & \textbf{17.1}                                                        \\ \bottomrule
\end{tabular}
\end{center}
\end{table*}